\documentclass[10pt,journal,compsoc]{IEEEtran}
\ifCLASSOPTIONcompsoc
  \usepackage[nocompress]{cite}
\else
  \usepackage{cite}
\fi
\ifCLASSINFOpdf
   \usepackage[pdftex]{graphicx}
\else
   \usepackage[dvips]{graphicx}
\fi
\usepackage{amsmath}
\ifCLASSOPTIONcompsoc
  \usepackage[caption=false,font=footnotesize,labelfont=sf,textfont=sf]{subfig}
\else
  \usepackage[caption=false,font=footnotesize]{subfig}
\fi

\usepackage[usenames, dvipsnames]{color}

\newcommand {\singbing}[1]{{\color{blue}\textbf{Sing Bing: }#1}\normalfont}

\begin{document}

%
\title{Hyperspectral Light Field Stereo Matching}
%
%
%
%

\author{Kang~Zhu, Yujia~Xue, Qiang~Fu, Sing~Bing~Kang, Xilin~Chen, and~Jingyi~Yu
	
\IEEEcompsocitemizethanks{
	\IEEEcompsocthanksitem K.~Zhu is with the School of Information Science and Technology, ShanghaiTech University, Shanghai, China, and the Key Laboratory of Intelligent
	Information Processing, Institute of Computing Technology, Chinese Academy
	of Sciences, Beijing, China.\protect\\E-mail: zhukang@shanghaitech.edu.cn
    \IEEEcompsocthanksitem Y.~Xue is with Plex-VR Inc., Shanghai, China.\protect\\E-mail: yujia.xue@plex-vr.com
    \IEEEcompsocthanksitem Q.~Fu is with the School of Information Science and Technology, ShanghaiTech University, Shanghai, China\protect\\E-mail: fuqiangx@outlook.com
    \IEEEcompsocthanksitem S.B.~Kang is with Microsoft Research, Redmond, WA.\protect\\    E-mail: sbkang@microsoft.com
    \IEEEcompsocthanksitem X.~Chen is with the Key Laboratory of Intelligent
    Information Processing, Institute of Computing Technology, Chinese Academy
    of Sciences, Beijing, China.\protect\\E-mail:xlchen@ict.ac.cn.
    \IEEEcompsocthanksitem J.~Yu is with the School of Information Science and Technology, ShanghaiTech University, Shanghai, China, and the Department of Computer and Information Sciences,
    University of Delaware, Newark, DE. \protect\\E-mail:yu@eecis.udel.edu}
\thanks{ }}

\IEEEtitleabstractindextext{
\begin{abstract}
In this paper, we describe how scene depth can be extracted using a hyperspectral light field capture (H-LF) system. Our H-LF system consists of a $5 \times 6$ array of cameras, with each camera sampling a different narrow band in the visible spectrum. 
There are two parts to extracting scene depth. The first part is our novel cross-spectral pairwise matching technique, which involves a new spectral-invariant feature descriptor and its companion matching metric we call bidirectional weighted normalized cross correlation (BWNCC). The second part, namely, H-LF stereo matching, uses a combination of spectral-dependent correspondence and defocus cues that rely on BWNCC. These two new cost terms are integrated into a Markov Random Field (MRF) for disparity estimation. Experiments on synthetic and real H-LF data show that our approach can produce high-quality disparity maps. We also show that these results can be used to produce the complete plenoptic cube in addition to synthesizing all-focus and defocused color images under different sensor spectral responses.
\end{abstract}

}

\maketitle
\IEEEdisplaynontitleabstractindextext
\IEEEpeerreviewmaketitle

\IEEEraisesectionheading{\section{Introduction}\label{sec:introduction}}
\begin{figure*}
	\centering
	\includegraphics[width=\linewidth]{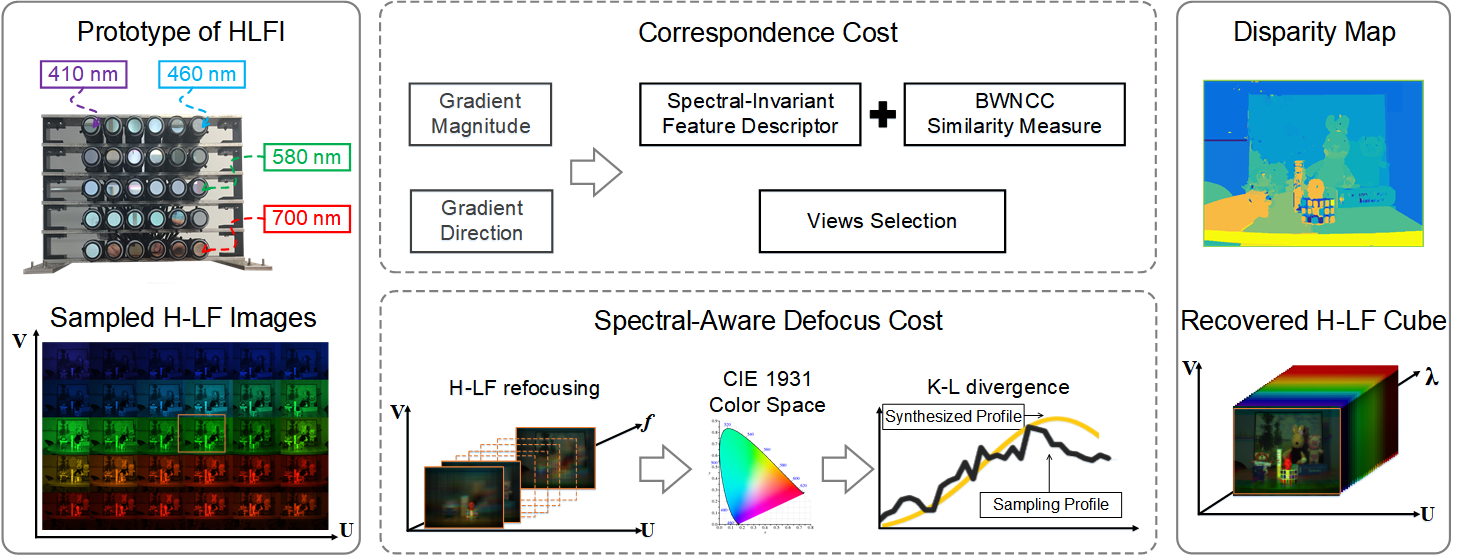}
	\caption{System overview. Our hyperspectral light field (H-LF) imager (HLFI, top left) consists of a $5\times6$ array of cameras, each with a narrow bandpass filter centered at a specific wavelength. The HLFI samples the visible spectrum from $410nm$ to $700nm$ with an $10nm$ interval. 
		We propose a new spectral-dependent H-LF stereo matching technique (middle), which involves novel correspondence cost (top) and spectral-aware defocus cost (bottom). The correspondence cost is based on a new spectral-invariant feature descriptor called BWNCC with local view selection.
		The generated disparity map (top right) can be used for complete plenoptic reconstruction (bottom right).
			} 
	\label{fig:pipeline}
\end{figure*}

\IEEEPARstart{T}{he} availability of commodity light field (LF) cameras such as Lytro~\cite{Ren2005Light} and Raytrix~\cite{Pappas2012Single} makes it easy to capture light fields. Dense stereo matching solutions have exploited unique properties, e.g., spatial and angular coherence~\cite{Tao2015coherence}, ray geometric constraints~\cite{Z_Yu2013LineAssistedLF,X_Guo2016RayspaceStitch,Tao2016LineConsistence}, focal symmetry~\cite{H_Lin2015FocalStack}, and defocus blurs~\cite{Tao2013Defocus}. In addition to 3D reconstruction, LF stereo matching can also address traditionally challenging problems, e.g., transparent object reconstruction~\cite{DingYuanyuan2011fluid}, saliency detection~\cite{N_Li2016Saliency} and scene classification~\cite{Wang2016Recognition}. 

Since an LF consists of densely sampled rays within a specific range of location and angle, it can be thought of as representing geometry and surface reflectance as well. However, the original plenoptic function \cite{Adelson91theplenoptic} includes an additional dimension of spectra, which has been largely ignored in most previous LF systems. What is required is hyperspectral imaging, which refers to the dense spectral sampling of a scene, as opposed to the regular RGB three-band sampling for color cameras. 

Holloway~et~al.~\cite{Holloway2015GAP} acquire a multispectral\footnote{In hyperspectral imaging, narrow band-pass filters are used. In the Holloway et al. work~\cite{Holloway2015GAP}, broadband filters are used instead.} LF using generalized assorted camera arrays. More recently, Xiong~et~al.~\cite{CVPR2017} adopts a hybrid sensing technique that combines an LF with a hyperspectral camera. These solutions assume small camera baselines for reliable image registration. In contrast, we present a wide baseline hyperspectral light field (H-LF) imaging technique based on novel cross-spectral LF stereo matching. 

Direct adoption of existing LF stereo matching solutions (e.g.,~\cite{Wanner2012Depth,Tao2013Defocus,Z_Yu2013LineAssistedLF,C_Chen2014Bilateral,Wang2015Occlusion,H_Lin2015FocalStack}) for H-LF would be ineffective, since images at different spectral bands may be very visually different. Instead, we introduce a new spectral-invariant feature descriptor and its companion matching metric (which we call bidirectional weighted normalized cross correlation or BWNCC). BWNCC measures gradient inconsistencies between cross-spectral images; it significantly outperforms other state-of-the-art metrics (e.g., sum of squared differences (SSD)~\cite{Shi-SSD}, normalized cross correlation (NCC)~\cite{NCC}, histogram of oriented gradient (HOG)~\cite{Dalal2005HOG}, scale-invariant feature transform (SIFT)~\cite{Lowe-SIFT}), in both robustness and accuracy. 

Our spectral-dependent H-LF stereo matching technique combines correspondence and defocus cues that are based on BWNCC. For visual coherency, we calculate the correspondence cost using local subsets of views, since views that are farther away may be less reliable. 
The entire H-LF is used to compute our new spectral-aware defocus cost. Previous approaches use color or intensity variance to measure focusness. However, for H-LF, the same 3D point will map to different intensities; as a result, such variance measures would be unreliable. We instead synthesize the RGB color from H-LF samples, then use the CIE Color Space to map the estimated hue of color to its spectral band. Consistency is then measured using the actual captured band as the focusness measure. Finally, we integrate the new correspondence and defocus costs with occlusion and smoothness terms in an energy function, and solve it as a Markov Random Field (MRF). 

We validate our approach on both synthetic and real H-LFs. To capture real H-LFs, we construct an H-LF camera array, with each camera equipped with a different narrow $10nm$-wide bandpass filter (Figure~\ref{fig:pipeline}). The union of all the cameras covers the visible spectrum from $410nm$ to $700nm$. The baseline is $36mm$, which is large enough that parallax for the scenes used would be significant. We show our H-LF stereo matching technique can produce high-quality disparity maps for both synthetic and real datasets. The disparity maps can be used to produce the complete plenoptic cube. These maps can also be used for image warping, which allows color image synthesis, hyperspectral refocusing, and emulation of different color sensors.

The contributions of this paper are:
\begin{itemize}
\item We designed a snapshot hyperspectral light field imager (HLFI) that samples only a subset of H-LFs, avoiding demosaicking artifacts. In principle, our HLFI can be expanded both in the spectral resolution (e.g., a $5nm$ interval or lower) and range (e.g., additional infrared and ultraviolet bands).
\item We propose a new spectral-invariant feature descriptor to effectively represent the visually-varying spectral images. We also propose a matching metric, BWNCC, to measure the similarity of multi-dimensional features. This feature descriptor and BWNCC are used in H-LF stereo matching.
\item We propose a novel spectral-dependent H-LF stereo matching technique that combines a local view selection strategy with spectral-aware defocus. We show that our matching technique produces high-quality disparity maps. 
\item We show three applications using our H-LF results: reconstruction of the complete plenoptic cube, generation of all-focus H-LF, and synthesis of defocused color images under different spectral profiles. We expect these applications would be useful for 3D reconstruction, object detection and identification, and material analysis.
\end{itemize}

The rest of this paper is organized as follows.
We review related work in feature descriptors, LF stereo matching, and multi-spectral imaging in Section~\ref{S_RELATED}.
Our design of HLFI is described in Section~\ref{S_SYS}.
Section~\ref{S_Stereo} details our feature descriptor and matching metric; these are used in our H-LF stereo matching technique (Section~\ref{S_SADE}).
The task of complete plenoptic cube reconstruction is described in Section~\ref{S_HLF_RES}.
Section~\ref{S_EXP} presents the experimental results and applications, with discussion of limitations in Section~\ref{S_DIS} and concluding remarks in Section~\ref{S_CON}.

\section{Related Work}\label{S_RELATED}

In this section, we review relevant approaches in the areas of multispectral imaging, feature descriptors, and light field stereo matching.

\subsection{Multispectral Imaging}

Spectral imaging has long been driven by the need of high quality remote sensing~\cite{goetz1985imaging}, with applications in agriculture, military, astronomy, surveillance, etc. (e.g.,~\cite{clevers1999the,curran2001imaging,bendor2008imaging,lockwood2007advanced}).
The commonly adopted techniques include coupling bandpass filters with spatial and temporal multiplexing to acquire both the spatial and spectral information. 
In satellite imaging, spectral-coded pushbroom cameras is capable of acquiring the full spectra~\cite{mouroulis2000pushbroom}. Tunable filters (e.g., LCTF, AOTF~\cite{gat2000imaging}) provide an alternative single camera solution. Such solutions require the camera be fixed under different shots and cannot provide scene parallax. More expensive snapshot imaging spectrometry involving diffraction grating, dispersing prism, multi-aperture spectral filter, Lyot filter or generalized Bayer filter (e.g.,~\cite{Su2015A,hagen2012snapshot}), requires extremely accurate calibration. 

Alternative approaches mostly rely on hybrid sensing, i.e., using sensors with different modalities. For example, Xiong et al.~\cite{CVPR2017} combine an LF camera with a hyperspectral camera to obtain the angular and spectral dimensions for recovery of the hyperspectral LF. 
Ye and Imai~\cite{Ye2015} describe a plenoptic multispectral camera whose microlens array has a spectrally-coded mask. Using a sparse representation, the spectral samples are used to reconstruct high resolution multispectral images. 
Closely to related to our work is that of generalized assorted cameras~\cite{Holloway2015GAP}, where a camera array is used for multispectral imaging. This system is a custom-built camera array (ProFUSION from PTGrey) that is modified by mounting broad band-pass filters. The camera baseline is rather small, and the filter being broad band-pass makes it easier to correspond images using existing color features. 

Our system uses a 2D array of monochrome cameras with narrow band-pass filters to avoid the demosaicking artifacts caused by the de-mulplexing procedure used in~\cite{Holloway2015GAP}. More importantly, the camera baselines in our system are significantly larger relative to scene depth; this allows more reliable depth estimation and enables synthetic refocusing. Finally, our system is extensible: in principle, more cameras can be added to increase the synthetic aperture (with wider extents) or spectral sampling resolution (with narrower band-pass filters).

\subsection{Feature Descriptors}
Feature descriptors (e.g.,~\cite{okutomi1993ssd}, \cite{Shi-SSD,Kolmogorov_ipol,NCC,ANCC,hirschmuller2008stereo,Holloway2015GAP,shen2014multi}) play a critical role in stereo matching and image registration. 
SSD (e.g.,~\cite{okutomi1993ssd},~\cite{Shi-SSD}) is widely used in stereo matching as a data cost (e.g., \cite{Kolmogorov_ipol}). 
NCC~\cite{NCC} is a highly popular as well for matching contrast-varying images. 
To take into account local radiometric variabilities, adaptive normalized cross correlation (ANCC)~\cite{ANCC} is introduced for matching.
Hirschmuller~\cite{hirschmuller2008stereo} uses mutual information (MI) with correlation-based method to resolve radiometric inconsistencies in images matching. 
Other matching features used include robust selective normalized cross correlation (RSNCC)~\cite{shen2014multi} for multi-modal and multispectral image registration, and cross-channel normalized gradient (CCNG)~\cite{Holloway2015GAP} for multispectral image registration.

However, in cross-spectral stereo matching, the crucial problem is the spectral difference. Techniques such as \cite{Shi-SSD, Kolmogorov_ipol, NCC} do not work well because of the intensity consistency assumption. Although radiometric inconsistencies that are handled in \cite{ANCC,hirschmuller2008stereo} are related to spectral difference, they have very different properties. Radiometric changes (e.g., caused by the varying exposure or lighting) mostly preserve the relative ordering of local scene point intensities. In contrast, in multispectral imaging, the relative ordering of local intensities can change arbitrarily, including order reversal. This is because different materials tend to have different responses at different wavelengths. Both \cite{shen2014multi,Holloway2015GAP} are applied to multispectral imaging. Unfortunately, in \cite{shen2014multi}, errors occur in regions with uncorrelated textures. Meanwhile, \cite{Holloway2015GAP} describes a technique that operations on broad band-pass RGB color channels; it is not expected to handle the single narrow band-pass channel images in our H-LF as well using \cite{ANCC}. 

\subsection{Light Field Stereo Matching}
Many stereo techniques have been proposed~\cite{scharstein2002taxonomy}, including local methods~\cite{NCC}, semi-global methods~\cite{hirschmuller2008stereo}, and global methods~\cite{Kolmogorov_ipol}. More recently, these techniques have been adapted for LFs. For exmaple, Wanner and Goldlucke\cite{Wanner2012Depth} extract the direction field in the Epipolar Image to estimate disparity. Yu~et~al.~\cite{Z_Yu2013LineAssistedLF} use geometric structures of 3D lines in ray space to improve depth with encoded line constraints. Tao~et~al.~\cite{Tao2013Defocus} introduce  the defocus cue combined with correspondence for depth estimation. Chen~et~al.~\cite{C_Chen2014Bilateral} propose a bilateral consistency metric to handle occluding and non-occluding pixels, while Lin~et~al.~\cite{H_Lin2015FocalStack} make use of the LF focal stack to recover depth. Wang~et~al.~\cite{Wang2015Occlusion} handle occlusion through edge detection. Again, these solutions cannot be directly applied for H-LF stereo matching; under spectral variations, regular data consistency measures (such as focusness) are no longer effective. Our spectral-dependent H-LF stereo matching technique addresses the cross-spectral inconsistency problem by using a spectral-invariant feature descriptor, applying local selection of views, and using spectral-aware defocus cues. We also handle occlusion in a manner similar to~\cite{Wang2015Occlusion}.

\section{Hyperspectral Light Field Imager (HLFI)}\label{S_SYS}
\begin{figure} 
	\centering
	\includegraphics[width=\linewidth]{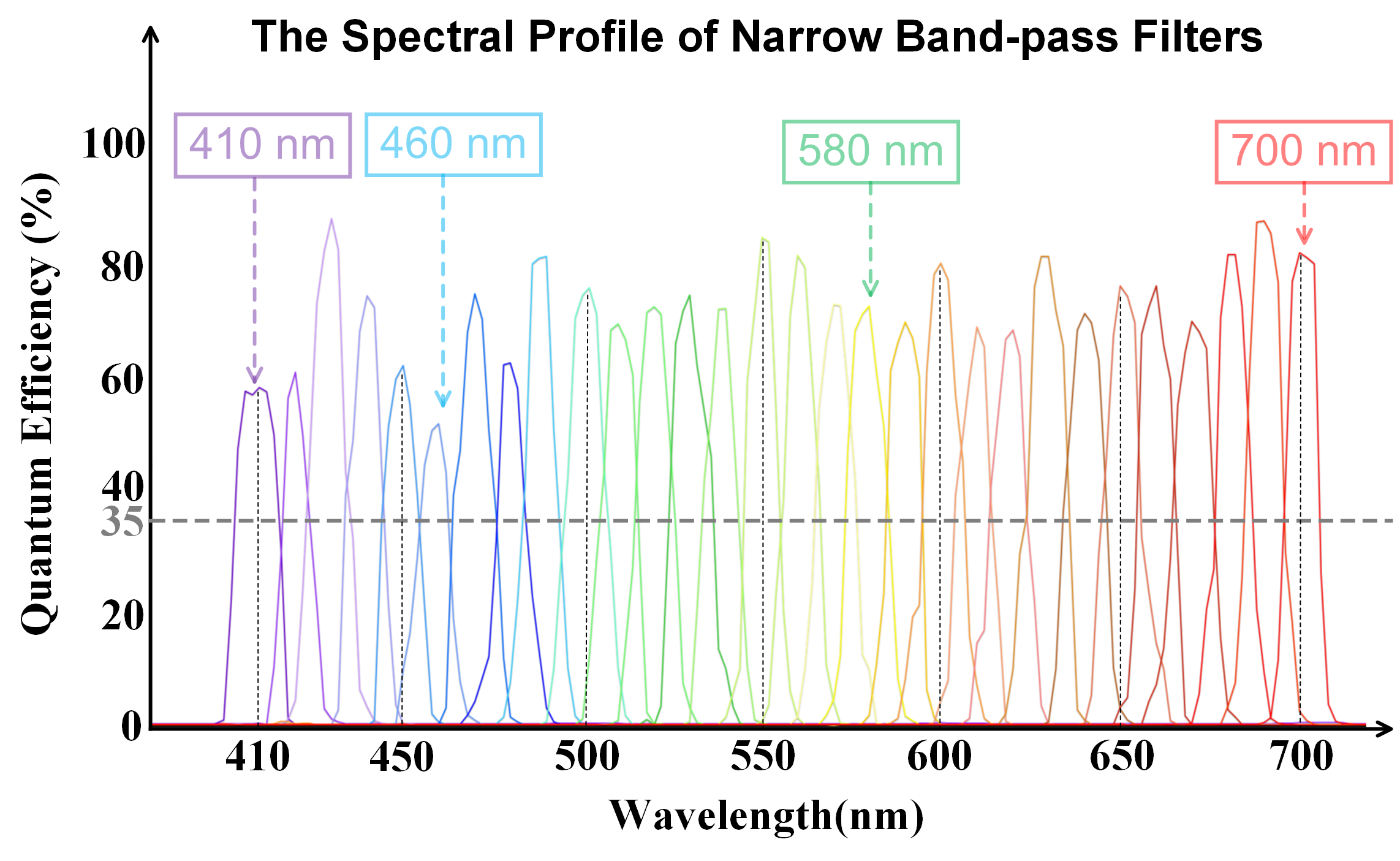}
	\caption{The spectral profile of narrow band-pass filters. In our setup, we mount 30 filters on camera array (Figure~\ref{fig:pipeline}). These filters sample the visible spectrum, centered from $410nm$ to $700nm$ with an $10nm$ interval as this figure shows. Each bandwidth is $10nm$~(i.e.,~$\pm 5nm$ about the central wavelength) with $\pm 2nm$ uncertainty. The overlaps occur near $35\%$ quantum efficiency with rapid drop-off.}
	\label{fig:filters}
\end{figure}

To simultaneously acquire spatial, angular, and spectral samples of the plenoptic function, we build a hyperspectral light field imager (HLFI). The left of Figure~\ref{fig:pipeline} shows our HLFI setup: we use an array of $5 \times 6$ monochrome cameras, each equipped with a narrow band-pass filter centered at a different wavelength. 
The spectral responses of the filters are shown in Figure~\ref{fig:filters}. These filters sample the visible spectrum, centered from $410nm$ to $700nm$ with an $10nm$ interval. The bandwidth of each filter is $10nm~(i.e.,~\pm 5nm)$ with $\pm 2nm$ uncertainty. Due to the uncertainty, neighboring filters have responses that overlap. Fortunately, the response drop-off for each narrow band-pass filter is steep. As shown in Figure~\ref{fig:filters}, the overlaps occur below $35\%$ quantum efficiency, where drop-off is rapid. We treat each filter response as a Dirac delta function $F_{\lambda_{i}}(\lambda)=\delta(\lambda - \lambda_{i})$, where $\lambda_i$ is the center wavelength.

To accommodate the extra spectral dimension, we modify the two-plane LF representation~\cite{Levoy1996Light,Ren2005Light} to $L(u, v, s, t, \lambda)$ for the sampled hyperspectral light field (H-LF). $(u, v)$ and $(s, t)$ represent the ray intersection with the aperture and sensor planes (respectively) at wavelength $\lambda$. The image $I(s,t,\lambda_{i})$ on $(s,t)$ corresponding to narrow band-pass spectral profile $F_{\lambda_{i}}(\lambda)$ centered at wavelength $\lambda_{i}$ is modeled as:  
\begin{equation}
\begin{aligned}\label{E_LF}
I(s,t,\lambda_{i})= \iiint & L(u,v,s,t,\lambda)A(u,v) C(\lambda) \\
&\cdot F_{\lambda_{i}}(\lambda) \cos^4{\theta} d\lambda du dv ,
\end{aligned}
\end{equation}
where $A(u,v)$ is the aperture function, $\theta$ is incident angle of the ray, and $C(\lambda)$ is the camera spectral response function.
We ignore $\cos^4{\theta}$ using the paraxial assumption. Equation~\ref{E_LF} simplifies to:
\begin{equation}
\label{E_MODEL}
\begin{aligned}
I(s,t,\lambda_{i})&= C(\lambda_{i}) \iint L(u,v,s,t,\lambda_{i})A(u,v)du dv\\
&= C(\lambda_{i})S(s, t, \lambda_{i}) ,
\end{aligned}
\end{equation}
where $S(\lambda_{i})$ is the latent radiance image at spectrum $\lambda_{i}$ while $C(\lambda_{i})$ is the spectral response function. 

\section{Two-View Spectral-Aware Matching}\label{S_Stereo}
In this section, we describe our new approach to matching two views $\mathcal{L}$ and $\mathcal{R}$ corresponding to narrow band spectra centered at two different wavelengths $\lambda_L$ and $\lambda_R$ (respectively). 

\subsection{Spectral-Invariant Feature Descriptor}\label{S_SIFD}
\begin{figure}
	\centering
	\includegraphics[width=\linewidth]{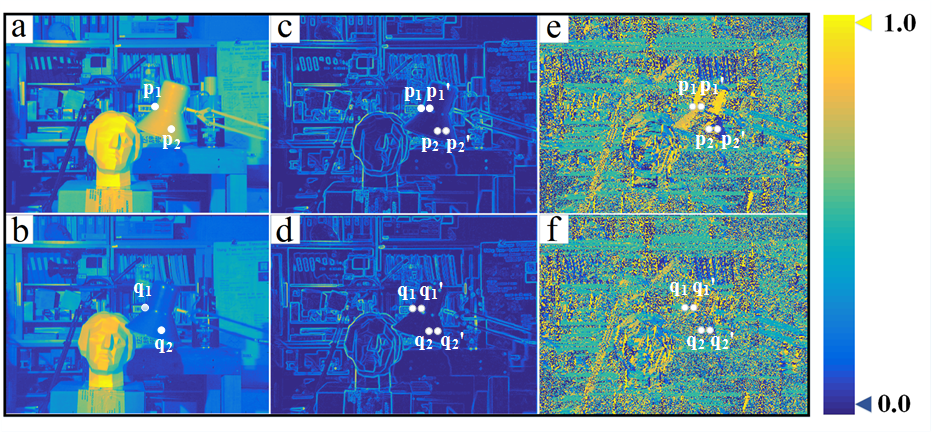}
	\caption
	{Cross-channel stereo imaging on the Tsukuba image pair.\hspace{10mm}
(a) and (b): Red channel of $\mathcal{L}$ and blue channel of $\mathcal{R}$, respectively. (c) and (d): respective gradient magnitudes.
(e) and (f): respective gradient directions. 
Section~\ref{S_SIFD} describes how we match boundary (e.g., $\mathbf{p}_1$-$\mathbf{q}_1$) and non-boundary (e.g., $\mathbf{p}_2$-$\mathbf{q}_2$) pixels. The pixels denoted with primes ($\mathbf{p}_1'$, etc.) are neighboring pixels.
	}\label{Gradient}
\end{figure}

Traditional measures for correspondence assume either brightness constancy or preservation of brightness ordering. As mentioned earlier, such measures (including direct gradient-based measures) fail because cross-spectral images violate these assumptions. Figure~\ref{Gradient} shows an example from the Middlebury dataset~\cite{scharstein2002taxonomy}. The red channel of $\mathcal{L}$ (Figure~\ref{Gradient}(a)) is markedly different from the blue channel of $\mathcal{R}$ (Figure~\ref{Gradient}(b)); for example, edge pixels around the lamp exhibit significant inconsistencies across the image pair. This demonstrates that we need to devise a new feature descriptor for cross-spectral images.  

We first eliminate the effect caused by the camera spectral response. From Equation~\ref{E_MODEL}, for two corresponding pixels $\mathbf{p}$ and $\mathbf{q}$ ($\mathbf{p},\mathbf{q}\in N^2$), we have $I_L({\mathbf{p}}) = C(\lambda_L)S_{\mathbf{p}}(\lambda_L)$ and $I_R({\mathbf{q}}) = C(\lambda_R)S_{\mathbf{q}}(\lambda_R)$. We normalize them to yield:
\begin{equation}
\label{E_L}
\left\{
\begin{aligned}
\widetilde{I}_L({\mathbf{p}}) &= \frac{I_L({\mathbf{p}})}{\bar{I}_L}=\frac{S_{\mathbf{p}}(\lambda_L)}{\bar{S}(\lambda_L)}\\
\widetilde{I}_R({\mathbf{q}}) &= \frac{I_R({\mathbf{q}})}{\bar{I}_R}=\frac{S_{\mathbf{q}}(\lambda_R)}{\bar{S}(\lambda_R)}
\end{aligned}
\right. ,
\end{equation}
where $\bar{I}_L$ and $\bar{I}_R$ are the mean intensities, and $\bar{S}(\lambda_L)$ and $\bar{S}(\lambda_R)$ are the average radiances in the corresponding views. For the remainder of the paper, we use $\widetilde{I}_L({\mathbf{p}})$ and $\widetilde{I}_R({\mathbf{q}})$ as inputs, eliminating the effect of the camera spectral response while still depending on the spectrum. We exploit the gradient of image as the feature descriptor. $M(\mathbf{p})$ and $\Theta(\mathbf{p})$ represent the magnitude and direction of the gradient at $\mathbf{p}$, respectively: $M({\mathbf{p}}) = \sqrt{\nabla_x \widetilde{I}(\mathbf{p})^2 + \nabla_y \widetilde{I}(\mathbf{p})^2}$ and $\Theta({\mathbf{p}}) = \mathbf{atan} \left( \nabla_y {\widetilde{I}(\mathbf{p})} / {\nabla_x \widetilde{I}(\mathbf{p})} \right)$. In Figure~\ref{Gradient}, (c) and (d) show the magnitudes of gradient for (a) and (b); (e) and (f) shows the directions of gradient for (a) and (b) quantized within $[0,\pi]$. 

We consider two cases, based on proximity to edges. \linebreak
\noindent
{\em Case 1}: Suppose corresponding pixels $\mathbf{p}, \mathbf{q}$ and their respective neighbors $\mathbf{p'}, \mathbf{q'}$ are all part of the same object (e.g., $\mathbf{p}_2, \mathbf{q}_2$ are adjacent to $\mathbf{p}'_2,\mathbf{q}'_2$, respectively, in Figure~\ref{Gradient}). Then, $|\widetilde{I}_L(\mathbf{p})-\widetilde{I}_L(\mathbf{p'})| \simeq |\widetilde{I}_R(\mathbf{q})-\widetilde{I}_R(\mathbf{q'})|$, implying that the gradient magnitude and direction should be approximately the same, i.e., $M_L({\mathbf{p}})\simeq M_R({\mathbf{q}})$ and $\Theta_L({\mathbf{p}})\simeq \Theta_R({\mathbf{q}})$. \linebreak 
\noindent
{\em Case 2}: Suppose the pixels lie near an edge (e.g., $\mathbf{p}_1, \mathbf{q}_1$ are adjacent to $\mathbf{p}'_1,\mathbf{q}'_1$, respectively, in Figure~\ref{Gradient}). The foreground and background correspond to objects with different spectral responses and the magnitude measure is no longer consistent. However, note that the gradient directions should still be similar. 

We design a feature descriptor that measures both edge and non-edge points. The non-edge features couple the gradient magnitude and direction histograms, whereas the edge features are an extension of HOG we call Overlapping HOG or O-HOG. Unlike traditional histograms, where every bin represent a separate range of values, in O-HOG, adjacent bins have overlapping values (i.e., they share some range of values). This is to more robustly handle view and spectral variations that exist in cross-spectral matching. By comparison, even a slight change in perspective or spectrum may lead to misalignment in regular HOG~\cite{Dalal2005HOG}.

To find correspondence, we first calculate the gradient magnitude and direction histograms (with $K_1$ and $K_2$ bins, respectively). 
Given a local window $\mathbf{U}({\mathbf{p}},w) \in \mathcal{N}^{w^2 \times 2}$ centered at $\mathbf{p}$ with size $w \times w$ for a stack of magnitude and direction images, we count weighted votes for bins in the magnitude histogram $\mathbf{h}_1(\mathbf{p},w,K_1)$ and direction histogram $\mathbf{h}_2(\mathbf{p},w,K_2)$. Specifically, the $k$-th  bin $b_i^{(k)}(\mathbf{p},w)$ of $\mathbf{h}_i (i = 1, 2 ;k\in[0,K_i -1))$  is aggregated as
\begin{equation} \label{E_NCHG}
b_i^{(k)}(\mathbf{p},w) = \frac{\sum \limits_{\mathbf{u}_t \in \mathbf{U}(\mathbf{p},w)} G(\mathbf{p},\mathbf{u}_t,\sigma_g) f(\mathbf{u}_t)}{\sum \limits_{j \in [0, K_i-1] }b_i^{(j)}} ,
\end{equation}
where  $G(\mathbf{p},\mathbf{u_t},\sigma_g) = \exp \left( {-||\mathbf{p}-\mathbf{u}_t||^2_2}/{2\sigma_g}^2 \right)$ is a spatial weight kernel, and
$f(\mathbf{u}_t)$ is a truncation function defined as
\begin{equation}
\label{eq:fut}
f(\mathbf{u}_t) = 
\left\{
\begin{aligned}
&1 	\quad 	\textnormal{if } Q(\mathbf{u}_t) \in [k(1-o)s,k(1-o)s+s)\\
&0	\quad	\textnormal{otherwise} 
\end{aligned}
\right. .
\end{equation}
Here $o$ is the overlapping portion between the neighboring bins and $s$ is the bin width. For $\mathbf{h}_1$, $Q(\mathbf{u}_t) = M(\mathbf{u}_t)$; for $\mathbf{h}_2$, $Q(\mathbf{u}_t) = \Theta(\mathbf{u}_t)$.
Similarly, for the O-HOG histogram $\mathbf{h}_3(\mathbf{p},w,K_3)$, the $k$-th bin $b_3^{(k)}(\mathbf{p},w)$ is computed as
\begin{equation} \label{E_NMHD}
b_3^{(k)}(\mathbf{p},w) = \frac{\sum \limits_{\mathbf{u}_t \in \mathbf{U}(\mathbf{p},w)} G(\mathbf{p},\mathbf{u}_t,\sigma_g) M(\mathbf{u}_t) f(\mathbf{u}_t)}{\sum \limits_{j \in [0, K_3-1] }b_3^{(j)}} .
\end{equation}
Note that for $\mathbf{h}_3$, $Q(\mathbf{u}_t) = \Theta(\mathbf{u}_t)$ in Equation~\ref{eq:fut}. We set $s=1/64$ and $o=1/16$ for both $\mathbf{h}_1$, $\mathbf{h}_2$ and $\mathbf{h}_3$. $K_1=K_2=K_3$, all round up to $68$.

We define descriptor $\mathbf{D_p} = \left[ \alpha_1  \mathbf{h}_1^T, \alpha_2  \mathbf{h}_2^T, \alpha_3  \mathbf{h}_3^T \right]^T$, with $\alpha_1$, $\alpha_2$, and $\alpha_3$ being weights. Recall that $\mathbf{h}_1$ and $\mathbf{h}_2$ represent non-edge points and $\mathbf{h}_3$ represents edge points. Since $M(\mathbf{p})$ is the edge strength of $\mathbf{p}$, we simply reuse $M(\mathbf{p})$ to get $\alpha_1=\alpha_2=\beta \exp({-M^2(\mathbf{p})}/{\sigma_w})$ and $\alpha_3=1-\alpha_1-\alpha_2$. In our work, $\beta=1/2$ and $\sigma_w=0.16$. 

\begin{figure}
	\centering
	\includegraphics[width=\linewidth]{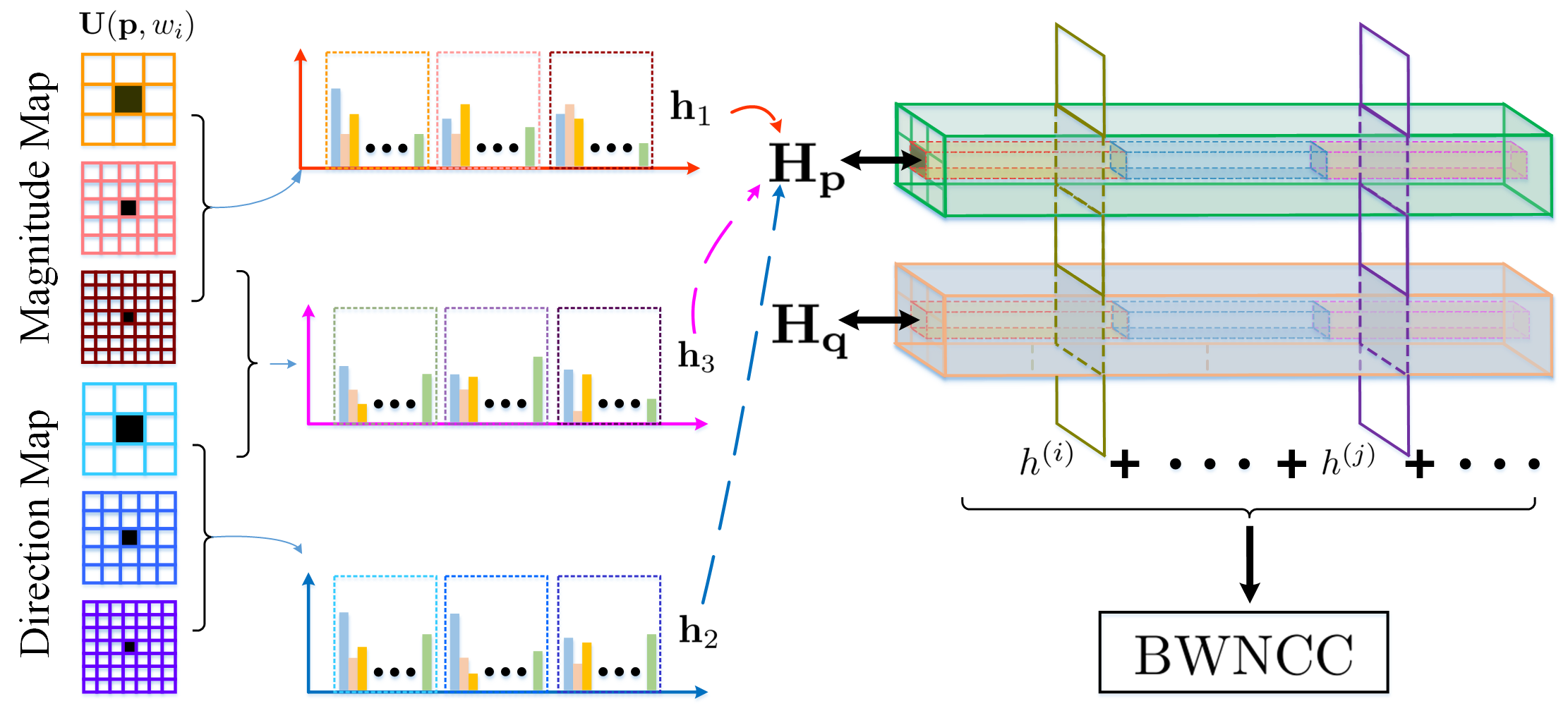}
	\caption{Our spectral-invariant feature descriptor $\mathbf{H}$ is based on weighted histograms for 3-level pyramids of the gradient magnitude and direction maps. $\mathbf{h}_1$ and $\mathbf{h}_2$ are the histograms for gradient magnitude and direction, while $\mathbf{h}_3$ represents O-HOG. 
} 
	\label{OHOG}
\end{figure}

For robustness, we build a 3-level pyramid structure with different patch widths $\mathbf{w}=[w_1,w_2,w_3]^T$ to obtain the final descriptor $\mathbf{H_p} = \left[ \mathbf{D_p^T}(w_1), \mathbf{D_p^T}(w_2), \mathbf{D_p^T}(w_3) \right]^T$ with $K$ elements, where $K=3(K_1+K_2+K_3)$. In all our experiments, $\mathbf{w}=[3,5,9]^T$.  Figure ~\ref{OHOG} shows the structure of our feature descriptor.

\subsection{Spectral-Invariant Similarity Metric}\label{S_NCC}
A popular similarity metric for stereo matching is the normalized cross correlation (NCC)~\cite{NCC}:
\begin{eqnarray}
\xi(I) = \frac{\sum \limits_{\mbox{\tiny $\begin{array}{c}\mathbf{u}_i \in \mathbf{U}_{L}\\ \mathbf{u}_j\in \mathbf{U}_{R}\end{array}$ }} 
	(I_L(\mathbf{u}_i)-\bar{I}_L)(I_R(\mathbf{u}_j)-\bar{I}_R)
}
{
	\sqrt{
		\sum \limits_{\mathbf{u}_i \in \mathbf{U}_{L}} (I_L(\mathbf{u}_i)-\bar{I}_L)^2
		\sum \limits_{\mathbf{u}_j \in \mathbf{U}_{R}} (I_R(\mathbf{u}_j)-\bar{I}_R)^2
	}
} ,
\end{eqnarray}
where $\bar{I}_L$ and $\bar{I}_R$ are the mean values of $\mathbf{U}_{L}(\mathbf{p},w)$ and $\mathbf{U}_{R}(\mathbf{q},w)$, respectively, in domain $I$ (e.g., intensity).

Unfortunately, NCC cannot be directly used to match multi-dimensional features. Note that each element $h^{(i)}$ in $\mathbf{H}$ is independent of any other element $h^{(j)}$ ($j \not =i$), and represents a unique attribute of $\mathbf{H}$ (as shown in Figure~\ref{OHOG}). We define our similarity metric as $\xi(\mathbf{H})=\sum \limits_{i=0}^{K-1} \omega_i \xi(h^{(i)})$, where $\omega_i$ is a similarity weight of $h^{(i)}$. In principle, we can simply use $h^{(i)}$ as $w_i$. In practice, for robustness to noise, we use the mean $\bar{h}^{(i)}$ instead of $h^{(i)}$ as weights.

Since $h_{\mathbf{p}}^{(i)}$ and $h_{\mathbf{q}}^{(i)}$ play equally important roles in computing $\xi{(\mathbf{H})}$, the final metric we use incorporates both, leading to the Bidirectional Weighted Normalized Cross Correlation (BWNCC). The forward component weighted by $\bar{h}_{\mathbf{p}}^{(i)}$ represents the similarity between $\mathbf{p}$ and $\mathbf{q}$, while the backward component weighted by $\bar{h}_{\mathbf{q}}^{(i)}$ represents the similarity between $\mathbf{q}$ and $\mathbf{p}$. BWNCC is thus defined as
\begin{equation}
\xi_{bwncc}(\mathbf{H}) = \sqrt{\sum \limits_{i=0}^{K-1} \xi{(h^{(i)})} \bar{h}_{\mathbf{p}}^{(i)} \sum \limits_{j=0}^{K-1} \xi{(h^{(j)})} \bar{h}_{\mathbf{q}}^{(j)}} .
\end{equation}

\section{H-LF Stereo Matching Scheme} \label{S_SADE}
Our new feature descriptor and metric enable more reliable feature selection and matching. Compared with binocular stereo, LF stereo matching has two different properties: use of many views and refocusing. When modeled as a disparity labeling problem, the correspondence cost makes use of the multiple views while defocus cost is based on refocusing (e.g.,~\cite{Tao2013Defocus,Z_Yu2013LineAssistedLF,H_Lin2015FocalStack,Wang2015Occlusion}). We denote $\Omega$ as all LF views $(s,t)$ and estimate the disparity map for the central view $(s_o,t_o)$. For simplicity, we use $I_{\mathbf{p}}(s,t)$ to represent $\widetilde{I}(u_{\mathbf{p}},v_{\mathbf{p}},s,t,\lambda_{(s,t)})$ in Equation~\ref{E_L}. 

The correspondence cost is typically cast as (\cite{Tao2013Defocus,Z_Yu2013LineAssistedLF,H_Lin2015FocalStack,Wang2015Occlusion}):
\begin{equation}
 C(\mathbf{p},f(\mathbf{p})) \propto \dfrac{1}{|\Omega|}  \sum_{(s,t)\in \Omega}|I_{\mathbf{p}}(s,t)-I_{\mathbf{p}}(s_o,t_o)|_2^2 .
\end{equation}  
For H-LF, we find a proper subset $\Omega^*$ ($\Omega^* \subseteq \Omega$) and use that in conjunction with our feature descriptor and metric to maximize spectral consistency. The defocus cost in \cite{Wang2015Occlusion} is based on the depth-from-defocus formulation for non-occlusion regions:
\begin{equation}
D(\mathbf{p},f(\mathbf{p})) \propto \nabla_{(x,y)}\bar{I}_{\mathbf{p}} .
\end{equation}
For occlusion regions the defocus cost is
\begin{equation}
D(\mathbf{p}, f(\mathbf{p}))\propto \dfrac{1}{|\Omega|}\sum_{(s,t)\in \Omega} |I_\mathbf{p}(s,t)-\bar{I}_{\mathbf{p}}|_2^2 ,
\end{equation}
where $\bar{I}_{\mathbf{p}}=1/|\Omega|\cdot \sum_{(s,t)\in \Omega} I_{\mathbf{p}}(s,t)$. However, direct use of the defocus measure in H-LF would fail due to spectral variance. We instead propose a new defocus cost based on hue-spectrum matching. After extracting two initial disparity maps $f^*_c$ (based on correspondence cost) and $f^*_d$ (based on defocus cost), we then impose regularization to generate the refined result $f^{\dagger}$. 

\subsection{Correspondence Cost} 
Recall that the correspondence cost measures similarity of corresponding pixels. For a hypothesized disparity $f({\mathbf{p}})$, we compute this cost using our spectral-invariant feature descriptor and BWNCC metric:
\begin{equation}
\label{E_C_Cost}
C(\mathbf{p},f(\mathbf{p}))=\frac{1}{|\Omega^*|}\sum_{(s,t)\in \Omega^*} - \log(\xi_{bwncc}(\mathbf{H})) .
\end{equation}

Instead of matching $\mathbf{p}$ in $(s_o,t_o)$ with pixel $\mathbf{q}$ across all LF views, we use only a subset of views $\Omega^*$ that share a coherent appearance (response). To do so, we first compute the arithmetic mean gradient magnitude over all $\mathbf{q}$. Next, we determine if the gradient magnitude of $\mathbf{p}$ is above or below the mean value. If it is above, then it is likely that $\mathbf{p}$ is an edge pixel; we use only pixels $\mathbf{q}$ in the H-LF views with a higher gradient magnitude. Similarly, if it is below, it is likely that $\mathbf{p}$ is a non-edge point, and we use only the ones with lower gradient magnitudes. 

In addition, we treat occluding and non-occluding pixels differently using the technique described in \cite{Wang2015Occlusion} to extract an initial disparity map $f^*_c$ based on correspondence cost. If $\mathbf{p}$ is non-occluding, $f^*_c(\mathbf{p})=\min_f \{C\}$. If  $\mathbf{p}$ is occluding, we partition $\Omega^*$ into occluder and occluded regions $\Omega^*_1$ and $\Omega^*_2$ (analogous to \cite{Wang2015Occlusion}), then compute $C_1$ and $C_2$ using Equation~\ref{E_C_Cost}. This yields $f^*_c (\mathbf{p})=\min_f\{C_1,C_2\}$.

\subsection{Defocus Cost} \label{S_DC}
A unique property in LF stereo matching is the availability of a synthetic focal stack, synthesized via LF rendering. Conceptually, if the disparity hypothesis is correct, the color variance over correspondences in all (non-occluding) views should be very small. If it is incorrect, the variance would be large, causing aliasing. In~\cite{Wang2015Occlusion}, the defocus cost measures the occlusion and non-occlusion regions separately in terms of color consistency. However, the traditional defocus cost cannot be used in our work because we cannot measure color consistency under different spectral responses. We adapted this cost to be spectral-aware. 

\begin{figure}
	\centering
	\includegraphics[width=\linewidth]{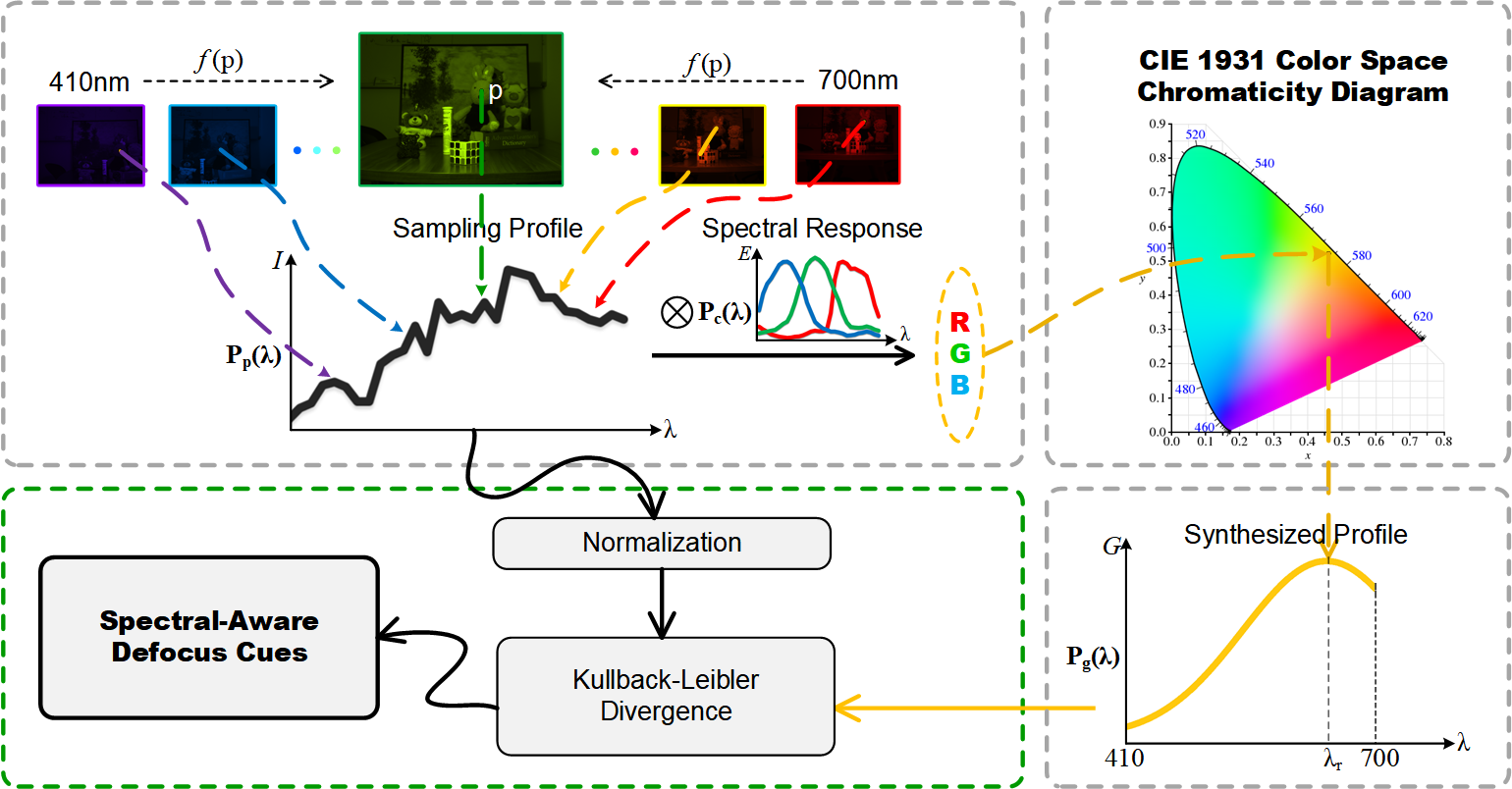}
	\caption{Spectral-aware defocus cue. Given a disparity hypothesis, we combine corresponding pixels from H-LF to form its spectral profile $\mathbf{P}_p(\lambda)$. Next, we use the camera (PTGrey FL3-U3-20E4C-C) spectral response curves $\mathbf{P}_c(\lambda)$ to map this profile to RGB color. We then convert the RGB color to its hypothesized wavelength $\lambda_r$ using the CIE 1931 Color Space. Finally, we match the observed profile with a Gaussian profile $\mathbf{P}_g(\lambda)$ centered at $\lambda_r$ via K-L divergence.}
	\label{fig:spectra_map}
\end{figure}

As Figure~\ref{fig:spectra_map} shows, given a hypothesized disparity $f(\mathbf{p})$, we estimate RGB color of $\mathbf{p}$ for a reference camera. To do this, we first form a spectral profile of $\mathbf{p}$ as $P_{\mathbf{p}}(\lambda)$ by indexing $\lambda_{(s,t)}$ using $I_{\mathbf{p}}(s,t)$ into respective views. Next, we use the spectral profile to synthesize its RGB value. In our experiments, we use the spectral response function of the PTGrey FL3-U3-20E4C-C camera (reference camera) as $\mathbf{P}_{c} (\lambda)=[P_r({\lambda}),P_g({\lambda}),P_b({\lambda})]^T$ and compute RGB values $\mathbf{V}=[R,G,B]^T$ by summing $P_{\mathbf{p}}(\lambda_{(s,t)}) \mathbf{P}_{c}(\lambda_{(s,t)})$  over the respective bandwidths:
\begin{equation}
 \mathbf{V}=\dfrac{\sum_{(s,t)\in \Omega} P_{\mathbf{p}}(\lambda_{(s,t)}) \mathbf{P}_{c}(\lambda_{(s,t)})}{\mathbf{P}_{c}(\lambda_{(s,t)})} .
\end{equation}

Finally, we map the RGB color back to spectra $\lambda_r$ by first converting it to hue before using a table to map hue to $\lambda_r$ based on CIE 1931 Color Space~\cite{Smith1931CIE}. 

If the disparity hypothesis is correct, $P_{\mathbf{p}}(\lambda)$ and the final RGB values estimation should be accurate. The captured spectra should then approximately form a Gaussian distribution centered at $\lambda_r$, with the probability density function
\begin{equation}
 P_{g}(\lambda)=\dfrac{1}{\sigma_d\sqrt{2\pi}} \cdot \exp{\left( -\dfrac{(\lambda-\lambda_r)^2}{2\sigma_d^2} \right)} .
\end{equation}
In our implementation, we use the special case of $\lambda_r=550nm$ (middle of $[410nm,700nm]$) to set $\sigma_d=96.5$. This is to ensure that $P_g(\lambda)$ have at least $30\%$ response in overlapping the visible spectrum throughout (especially in the corner cases of $\lambda = 400nm$ and $\lambda = 700nm$).

We subsequently normalize $P_{\mathbf{p}}(\lambda)$ to $P_{\mathbf{p}}^*(\lambda)=P_{\mathbf{p}}(\lambda)/\sum_{(s,t)\in \Omega}P_{\mathbf{p}}(\lambda_{(s,t)})$, and measure the Kullback--Leibler divergence~\cite{kullback1951,MacKay2003Information} from $P_{\mathbf{p}}^*(\lambda)$ to $P_g(\lambda)$. This results in our defocus cost
\begin{equation}
\label{E_D_Cost}
D(\mathbf{p},f(\mathbf{p}))=\sum_{(s,t)\in \Omega} P_g(\lambda_{(s,t)})\log{\frac{P_g(\lambda_{(s,t)})}{P_{\mathbf{p}}^*(\lambda_{(s,t)})}} .
\end{equation}
Finally, we have $f^*_d(\mathbf{p})=\min_f\{D\}$.

\subsection{Regularization}
The energy function for disparity hypothesis $f$ that is typically used in an MRF is (\cite{Tao2013Defocus,Wang2015Occlusion})
\begin{equation}
\label{E_MRF}
E(f)=E_{unary}(f)+E_{binary}(f) .
\end{equation}
We adopt the binary term similar to Wang~et~al.~\cite{Wang2015Occlusion} for smoothness and to handle occlusion. The major difference is that we use spectral-aware defocus cues (described in Section~\ref{S_DC}).

Our unary term is defined as
\begin{equation}\label{E_UN}
\begin{aligned}
E_{unary}(f)=\sum_{\mathbf{p}} &\gamma_c |C(f(\mathbf{p}))-C(f^*_c(\mathbf{p}))|\\&+|D(f(\mathbf{p}))-D(f^*_d(\mathbf{p}))| ,
\end{aligned}
\end{equation}
where $\gamma_c$ adjusts the weight between defocus and correspondence cost. (Its value is $0.45$ for synthetic data and $0.6$ for real data.)
Minimizing this function yields the desired disparity map $f^{\dagger}$.

\section{Plenoptic Cube Completion}\label{S_HLF_RES}
The results of our H-LF stereo matching technique can be used to complete the missing dimensions. The direct approach would be to use the disparity map of the central view to warp images to propagate the missing information. 
The problem, however, is that the warped images will contain holes due to occlusion. While it is possible to perform independent pairwise stereo matching between all views, this approach does not fully exploit the properties of LFs. We instead present a technique for cross-spectral joint binocular stereo.

\subsection{Disparity Initialization} \label{S_DI}
We first warp the disparity map of the central view, $f^{\dagger}_{(s_0,t_0)}$ (using the technique described in Section~\ref{S_SADE}) to individual LF views as their initial disparities:
\begin{equation}
\begin{aligned}
f^*_{(s,t)}(u+d(s-s_0),v+d(t-t_0))=d ,
\end{aligned}
\end{equation}
where $d=f_{(s_0,t_0)}^{\dagger}(u,v)$. At this point, at each view, $f_{(s,t)}^*(u,v)$ is an incomplete disparity map. There are pixels with invalid depth due to occlusion, being outside the field-of-view of the central view, and mismatches. However, regions that are valid can be used to guide and refine correspondences between cross-spectral image pairs.

\subsection{Disparity Estimate}
Using our BWNCC metric described in Section~\ref{S_Stereo}, we extract the disparity map for an image pair using Graph Cuts~\cite{Kolmogorov_ipol}. The energy function in~\cite{Kolmogorov_ipol} is
\begin{equation}
E(f)=E_{data}(f)+E_{occlu}(f)+E_{smooth}(f)+E_{unique}(f) .
\end{equation}
$E_{data}(f)$ is the data term that calculates similarity between corresponding pixels: 
\begin{equation}
\begin{aligned}
E_{data}(f)=\sum_{\mathbf{p}}|C(f(\mathbf{p}))-C(f^*(\mathbf{p}))| ,
\end{aligned}
\end{equation}
where $C(f(\mathbf{p}))$ is defined in Equation \ref{E_C_Cost} but applied to two views (i.e., left and right or top and bottom).
$E_{occlu}(f)$ is the occlusion term to minimize the number of occluded pixels while $E_{smooth}(f)$ is the smoothness term that favors piecewise constant maps. $E_{unique}$ enforces uniqueness of disparities between image pairs. The last three terms are same as those in~\cite{Kolmogorov_ipol}.

The disparity maps for all image pairs (with vertical or horizontal neighbors) are computed using Graph Cuts~\cite{Kolmogorov_ipol}. These disparity maps are then merged to produce a single one denoted as $f_{(s,t)}$. 

\subsection{Disparity Refinement}
As mentioned in Section~\ref{S_DI}, $f^*_{(s,t)}$ has regions of invalid depth. Furthermore, $f_{(s,t)}$ is likely to have unreliable depths due to occlusion in neighboring views.

Park et al.~\cite{park2014high} propose an optimization technique using RGB-D images to acquire a high-quality depth map. This technique uses confidence weighting in terms of color similarities, segmentation, and edge saliency. We use a similar approach to refine $f_{(s,t)}$, with the difference being the confidence weighting is adapted to our single channel spectral images. This results in the improved disparity map $f^{\dagger}_{(s,t)}$ for each view.

\subsection{Image Registration}
We use $f^{\dagger}_{(s,t)}$ to warp images. First, all pixels $\mathbf{p}$ on left (or top) view are mapped to $\mathbf{q}$ on right (or bottom) view. We then register all images currently on right (or bottom) to left (or top) view. This is iterated for all neighboring pairs until the plenoptic cube is completed, i.e., when all the missing spectra are propagated across all the views. For an H-LF imager with $M \times N$ views, the number of hyperspectral images in the completed plenoptic cube is $M \times N \times MN$, with each view having $MN$ images corresponding to $MN$ different spectra.

\section{Experiments and Applications}\label{S_EXP}
In this section, we report the results of our technique and how they compare with competing state-of-the-art.
We also describe two applications (namely, color sensor emulation and H-LF refocusing) that are made possible using the depth information generated using our technique.

\subsection{Experimental Setup} 
Our prototype HLFI consists of a $5 \times 6$ monochrome camera array (Figure~\ref{fig:pipeline}). The cameras are MER-132-30GM from Daheng ImaVision, with a resolution of $1292\times964$; they are synchronized via GenLock. The lens are M0814-MP2 from Computar, with a focal length of $8mm$. 
We mount 30 narrow bandpass filters (from Rayan Technology) on cameras centered wavelengths between $410nm$ to $700nm$ at a $10nm$ interval. Data collection and processing  are done on a Lenovo ThinkStation P500 with a Intel(R) Xeon(R) 4-core CPU E5-1630 running at 3.70GHz. 

We calibrate our cameras using Zhang's algorithm~\cite{zhang2000flexible} to extract the intrinsic and extrinsic parameters. Since we use a black-and-white checkerboard and all the filters are within the visible spectra, the calibration images have sufficient contrast for corner detection. Once the cameras are calibrated, the views are rectified to simplify stereo matching. As mentioned in Section~\ref{S_SIFD}, throughout all our experiments, we set $s=1/64$, $o=1/16$, and $\mathbf{w}=[3,5,9]^T$ to generate our hierarchical feature descriptor $\mathbf{H}$. 

\subsection{Validation for Feature Descriptor with BWNCC} 
We compared results of pairwise stereo matching using Graph Cuts~\cite{Kolmogorov_ipol} with SSD~\cite{Shi-SSD}, NCC~\cite{NCC}, and the recent RSNCC~\cite{shen2014multi} measures against those for our spectral-invariant descriptor with BWNCC measure. 

\begin{figure*}
	\centering
	\includegraphics[width=1.05\linewidth]{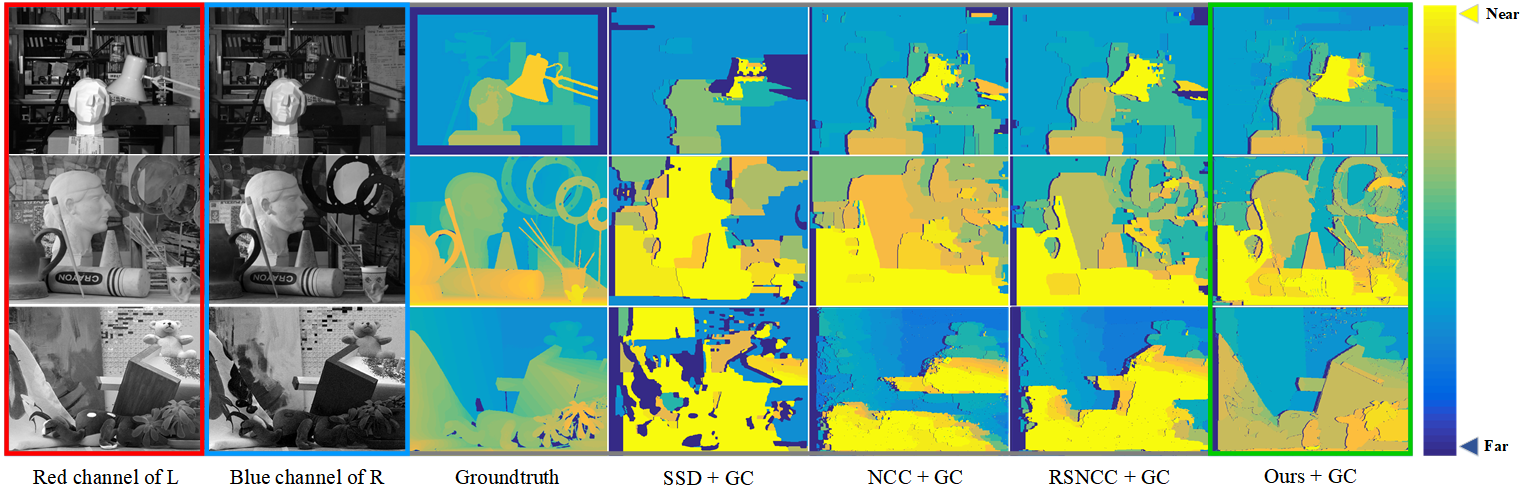}
	\caption{Cross-channel stereo matching results for three Middlebury datasets. From left to right: red channel of the left image, blue channel of right image,
		ground truth disparity map, estimated disparity map by SSD, NCC, RSNCC methods with graph cuts, and our proposed feature descriptor with BWNCC metric. Our method can estimate much better disparity maps compared with these state-of-the-art methods.
	}\label{fig:feature_syn}
\end{figure*}

We first ran experiments involving synthetic data adapted from the Middlebury stereo vision datasets~\cite{scharstein2002taxonomy}. To emulate spectrum inconsistency, we treat the red channel of $\mathcal{L}$ and the blue channel of $\mathcal{R}$ as the pseudo cross-spectral pair. Figure~\ref{fig:feature_syn} compares the visual quality of the results using different methods and Table~\ref{tbl:feature_error} shows the quantitative comparisons in terms of bad5.0 (percentage of ``bad'' pixels whose error is greater than 5 pixels~\cite{scharstein2002taxonomy}). Our approach significantly reduces error in stereo matching. 

\begin{figure*}
	\centering
	\includegraphics[width=1.05\linewidth]{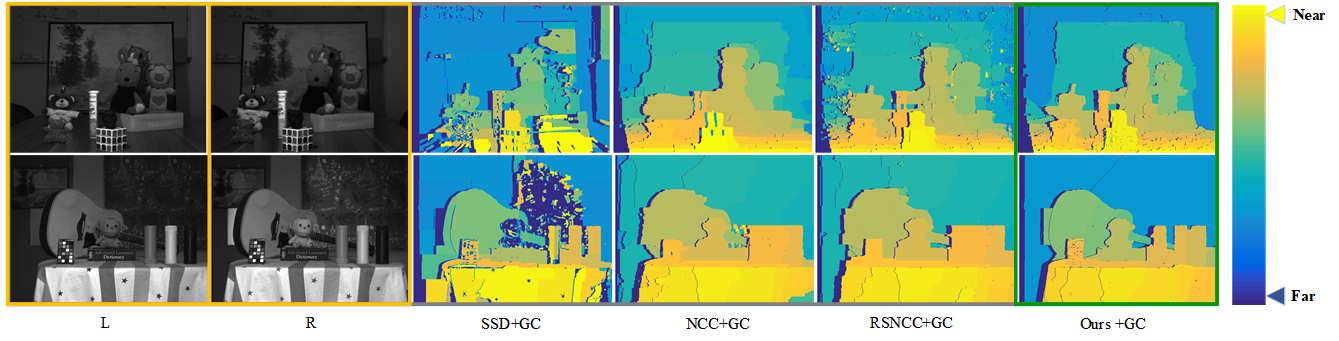}
	\caption{Cross-spectral stereo matching results on real scenes captured using our HLFI. The first and second columns are left and right images captured by two adjacent cameras at different spectra. The other columns show extracted disparity maps using SSD, NCC, RSNCC and our technique. Qualitatively, our results outperform the other competing techniques.}
	\label{fig:feature_real}
\end{figure*}

We also ran experiments on datasets captured using our HLFI on real scenes, again comparing our method with other competing techniques. Figure~\ref{fig:feature_real} shows results for two scenes. Visually, our approach outperforms the other techniques; for example, as can be seen at the bottom row, our technique is able to recover the guitar edge where other techniques fail due to spectral inconsistencies.

\begin{table}[h]
	\centering
	\caption{Comparison of bad5.0 error metric (smaller values are better).}
	\begin{tabular}{|c|c|c|c|}
		\hline
		&Tsukuba		&Art			&Teddy	\\
		\hline 
		\textbf{Ours}	&\textbf{3.14}	&\textbf{10.27}	&\textbf{7.01}	\\
		\hline
		RSNCC			&5.27			&16.64			&11.02	\\
		\hline
		NCC				&6.09			&18.31			&15.68	\\
		\hline
		SSD				&11.18			&28.19			&43.53	\\
		\hline
	\end{tabular}
	\label{tbl:feature_error}
\end{table}

\subsection{H-LF Stereo Matching Results} 
In one set of experiments, we generate synthetic H-LF scenes from regular LFs used in Wanner~et~al.~\cite{Wanner2012Depth}. For each scene, we choose $5 \times 6$ views with uniform baseline. For each view, we add a synthetic tunable filter. Finally, we render $5 \times 6$ spectral images from original images by adjusting the filter transmittance in the RGB channels and converting them to gray scale. Because this synthesized spectral profile is different from that for our HLFI,  we choose different values of $\gamma_c$ in Equation~\ref{E_UN} ($0.45$ for synthetic data and $0.6$ for real data).

\begin{figure*}
	\centering
    	\includegraphics[width=\linewidth]{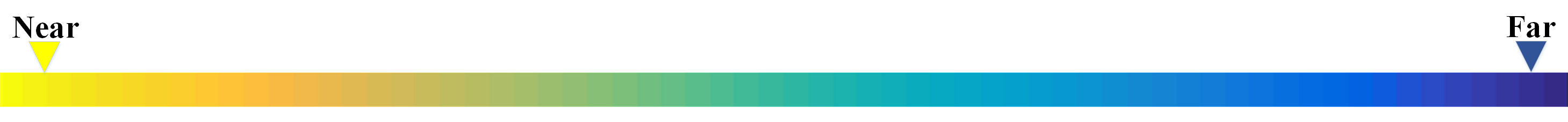}
	 \includegraphics[width=\linewidth]{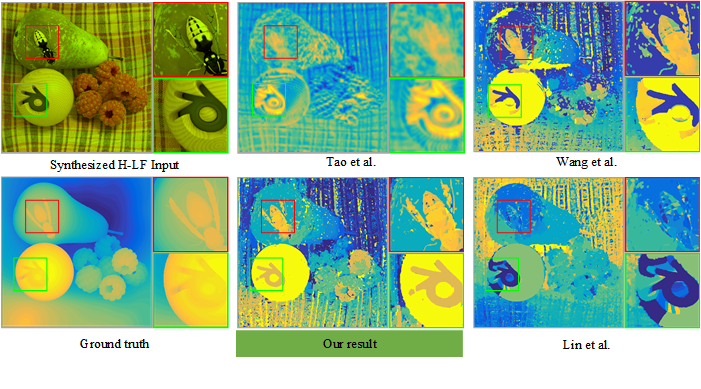}

	 \includegraphics[width=\linewidth]{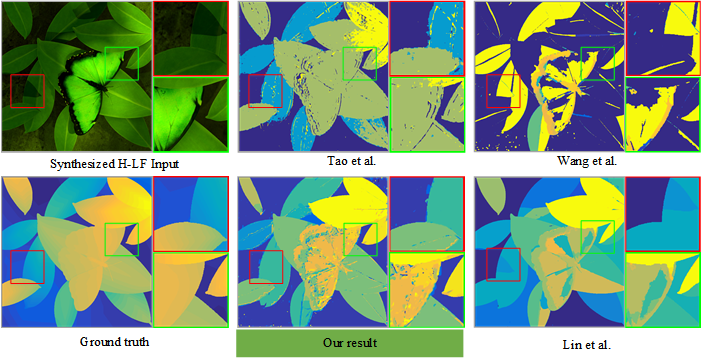}
	
	\caption{H-LF results for two synthetic scenes from Wanner~et~al.~\cite{Wanner2012Depth}, with each view having a different spectral response. We show our result as well as those of previous LF stereo matching methods (Tao~et~al.~\cite{Tao2013Defocus}, Lin~et~al.~\cite{H_Lin2015FocalStack}, and Wang~et~al.~\cite{Wang2015Occlusion}).
    The two close-ups show the relative quality of our result.}
	\label{fig:stereo_syn}
\end{figure*}

In another set of experiments, we compare our H-LF stereo matching results with techniques by Tao~et~al.~\cite{Tao2013Defocus}, Lin~et~al.~\cite{H_Lin2015FocalStack} and Wang~et~al.~\cite{Wang2015Occlusion}, on synthetic and real data.
Figure~\ref{fig:stereo_syn} compares the disparity maps on the synthetic dataset. The close-up regions (red and green boxes) show how well our technique works compared to the others. As Table~\ref{tbl:stereo_error} shows, our technique has the lowest RMSE.

\begin{figure*}
	\centering
	\includegraphics[width=0.95\linewidth]{color_bar.png}
	\includegraphics[width=0.95\linewidth]{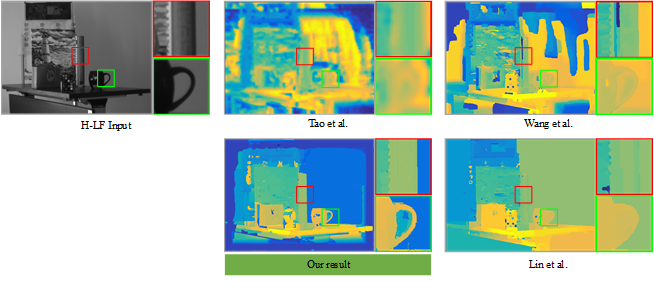}
	
	\includegraphics[width=0.95\linewidth]{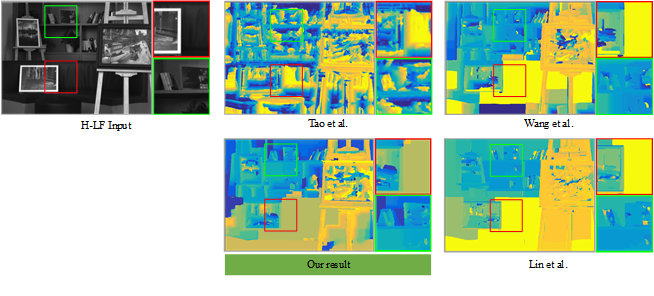}
	
		\includegraphics[width=0.95\linewidth]{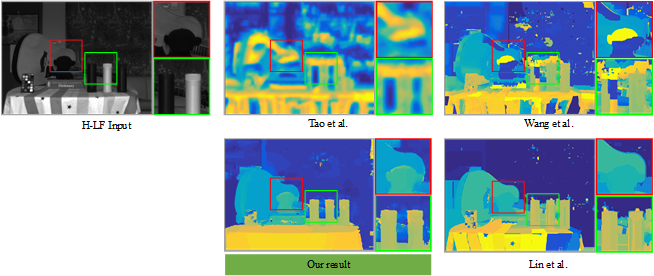}
	
	\caption{H-LF stereo matching results for three real scenes captured by our HLFI. 
    We show our results as well as those of previous LF stereo matching methods (Tao~et~al.~\cite{Tao2013Defocus}, Lin~et~al.~\cite{H_Lin2015FocalStack}, and Wang~et~al.~\cite{Wang2015Occlusion}).
	The two close-ups show how well our technique can recover scene detail.}
	\label{fig:stereo_real}
\end{figure*}

Figure~\ref{fig:stereo_real} shows H-LF stereo matching results for three real scenes. The overall visual quality of our results is better than that for the other competing techniques. In particular, our technique is better able to handle scene detail; see, for example, the mug in the top scene and guitar's neck in the bottom scene.

These results show our approach outperforms the state-of-the art in visual quality, accuracy, and robustness on both synthetic and real data. They validate our design decisions on handling cross-spectral variation. 

\begin{table}[h]
	\centering
	\caption{Comparison of RMSE (smaller values are better) for H-LF stereo matching on synthetic data shown in Figure~\ref{fig:stereo_syn}.}
	\begin{tabular}{|c|c|c|c|c|}
		\hline
		&Tao et al.		&Wang et al.			&Lin et al. &\textbf{Ours}	\\
		\hline
		Top scene&0.2052			&0.2970			&0.4285			&\textbf{0.1958}	\\
		\hline
		Bottom scene&0.4393				&0.3690			&0.2785			&\textbf{0.2266}	\\
		\hline
	\end{tabular}
	\label{tbl:stereo_error}
\end{table}

\subsection{H-LF Reconstruction Results}
\begin{figure}
	\centering
	\includegraphics[width=\linewidth]{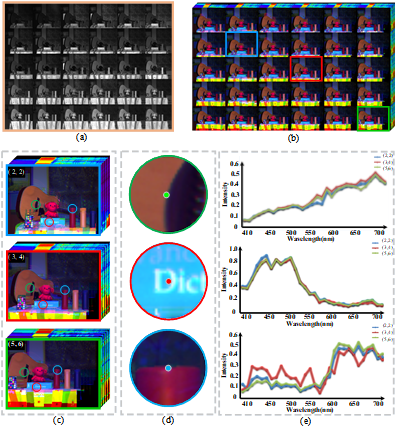}
	\caption{H-LF reconstruction results for a real scene. (a) Raw data acquired by our HLFI, (b) completed plenoptic cube, (c) reconstructed hyperspectral datacubes at viewpoints~(2, 2),~(3, 4),~and~(5, 6), (d) close-ups of representative boundary and textured areas, (e) spectral profiles of three scene points: a point on the guitar, a cyan point surrounded by white letters, and a depth boundary.}
	\label{fig:recons_real}	
\end{figure}

In another experiment, we use our HLFI to capture a room scene, processed the data using our technique, and completed its plenoptic cube representation. Results are shown in Figure~\ref{fig:recons_real}. The raw data are shown in Figure~\ref{fig:recons_real}(a); the scene has colorful objects made with different materials and placed at different depths.
Figure~\ref{fig:recons_real}(b) shows the completed plenoptic cube.
Reconstructed hyperspectral datacubes at viewpoints~(2, 2),~(3, 4),~and~(5, 6) are shown in Figure~\ref{fig:recons_real}(c).
Selected close-ups in Figure~\ref{fig:recons_real}(d) demonstrate that our technique can robustly align occlusion and texture boundaries under spectral variation.
Figure~\ref{fig:recons_real}(e) shows the spectral profiles of three scene points:
a point on the guitar, a cyan point surrounded by white letters, and a depth boundary. These results show that our reconstruction scheme can robustly align occlusion and texture boundaries under spectral variations and recover high fidelity H-LFs. 

\subsection{Applications: Color Sensor Emulation and H-LF Refocusing}

We can use the recovered H-LF data to emulate a synthetic camera with a specific spectral profile. This allows us to reproduce color images unique to that camera.

\begin{figure}
	\centering
	\includegraphics[width=\linewidth]{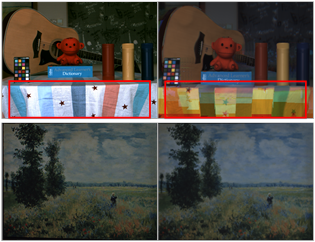}
	\caption{Comparison of real and synthetic color images. 
    Left: real images captured by a PTGrey FL3-U3-20E4C-C camera (the profile is same as that shown in Figure~\ref{fig:spectra_map}). 
    Right: synthesized images using our acquired H-LF and the camera profile.
    Given left images as references, PSNR of right image on top pair is 22.6 after removing the region of red box and alignment, whereas PSNR is 23.1 on bottom pair.
    }
	\label{fig:color_reproduce}
\end{figure}

Figure~\ref{fig:color_reproduce} shows two pairs of real images captured by PTGrey FL3-U3-20E4C-C alongside our synthesized color images (whose original spectral profile is shown in Figure~\ref{fig:spectra_map}).
The top pair includes original images (without cropping and alignment) of one scene. The red boxes show the incorrect color of the table cloth in the synthesized one (right), which is caused by missing spectra due to limited field of view. Notice that the top rows of Figure~\ref{fig:recons_real}(b) do not include most of the table cloth; as a result, no information on a specific range of the spectrum is available for propagation, causing incorrect color synthesis. After removing the region of red box and aligning images, we get PSNR of right image is $22.6$, given the left image as reference. The bottom pair includes cropped and aligned images of another scene, and PSNR of right image is $23.1$. Both the images and PSNR values show that our synthesized color images are reasonable reproductions of the actual versions.

\begin{figure}
	\centering
	\includegraphics[width=\linewidth]{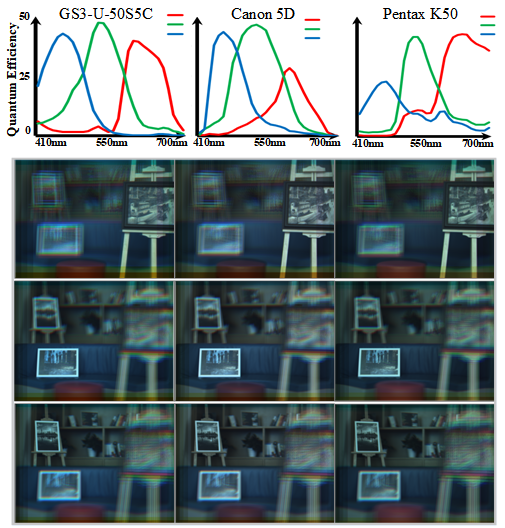}
	\caption{H-LF refocusing results.  
    Top: spectral profiles of three cameras. 
    Bottom: synthetic refocusing results at different depths for the three profiles.
    Results in different rows are at different depths (near, middle and far). Results in different columns are synthesized from different profiles respectively. }

	\label{fig:refocus}
\end{figure}

Figure~\ref{fig:refocus} shows results of synthetic refocusing for different spectral profiles. These results demonstrate that our dynamic H-LF refocusing is different from regular LF refocusing; it can focus at any depth layer at any sampled spectrum. Note that the banding artifacts are due to the discrete view sampling of our HLFI.

\section{Discussion}\label{S_DIS}

Because our HLFI is fundamentally a multi-view camera system, it has the same issues associated with length of baseline versus accuracy and ease of correspondence. Our HLFI has two main problems that are specific to multi-spectral matching. The first is the computational complexity of our feature descriptor and metric.  In order to acquire accurate depth, we need to consider both edge and non-edge regions hierarchically, and compute the distance using descriptors over local patches at different levels. These operations are more computationally expensive compared to traditional methods. Our GPU implementation produces depth results in about 2 minutes for datasets shown in Figure~\ref{fig:stereo_real} (each dataset has a $5 \times 6$ array of images, with each image having a resolution of $1200 \times 900$). 

Another problem is the incomplete spectral reconstruction due to missing views.
Each camera samples a narrow band of the visible spectrum and a different view of the scene. As a result, different parts of the scene would visible to a different subset of cameras in the HLFI. This results in incomplete propagation of the missing spectra, as can be seen in Figures~\ref{fig:recons_real} and~\ref{fig:color_reproduce}. More specifically, the table cloth has incorrect colors because cameras at the top few rows are not able to capture its appearance, resulting in absence of certain spectral bands.

There is also the interesting issue of filter arrangement. Currently, the filter wavelengths in our HLFI are arranged in raster order. As a result, as can be seen on the left of Figure~\ref{fig:pipeline}, horizontal neighbors are much more similar in appearance than vertical neighbors. This arrangement has implications on the H-LF stereo matching and reconstruction. There may be a better way of arranging these filters so as to reduce appearance changes in both vertical and horizontal directions. While it is possible to redesign with different cameras having the same filters, this reduces the spectral sampling density (for the same number of cameras and overall visible spectral extent).

\section{Concluding Remarks}\label{S_CON}

We have presented a hyperspectral light field (H-LF) stereo matching technique. Our approach is based on a new robust spectral-invariant feature descriptor to address intensity inconsistency across different spectra and a novel cross-spectral multi-view stereo matching algorithm. For increased robustness in matching, we show how to perform view selection in addition to measuring focusness in an H-LF. We have conducted comprehensive experiments by constructing an H-LF camera array to validate our claims. Finally, we show how our results can be used for plenoptic cube completion, emulation of cameras with known spectral profiles, and spectral refocusing.
 
An immediate future direction is to capture and process H-LF video. This will require temporal regularization techniques, in addition to requiring efficient compression to save bandwidth. In our current setup, the band-pass filters were sequentially assigned to the cameras, i.e., the neighboring cameras will have close spectral responses. The advantage of this setup is that we can more reliably conduct stereo matching and hence warping between adjacent images. Despite this, the baseline of the cameras cannot be too large, because we still require good visual overlap between images for effective spectral propagation. It would be interesting to investigate other camera designs with different spectral distributions to handle current limitations.

\bibliographystyle{IEEEtran}
\bibliography{IEEEabrv,reference}




\end{document}